\pdfoutput=1

\documentclass[11pt]{article}
\usepackage{makecell}
\usepackage{amsmath}
\usepackage{amssymb,amsthm}
\usepackage{newunicodechar}
\usepackage{placeins}
\usepackage{algpseudocode}
\usepackage{algorithm}
\usepackage[utf8]{inputenc}
\usepackage{textcomp}

\newunicodechar{ }{\,}
\usepackage[final]{acl}
\usepackage{times}
\usepackage{latexsym}

\usepackage[T1]{fontenc}

\usepackage[utf8]{inputenc}

\usepackage{microtype}

\usepackage{inconsolata}

\usepackage{graphicx}

\title{Dynamic Retriever for In-Context Knowledge Editing via Policy Optimization}

\author{
  Mahmud Wasif Nafee$^{1,2}$\thanks{~Equal contribution.} \quad
  Maiqi Jiang$^{3}$\footnotemark[1] \quad
  Haipeng Chen$^{3}$ \quad
  Yanfu Zhang$^{3}$\thanks{~Corresponding author.} \\
  $^{1}$Rensselaer Polytechnic Institute, Troy, NY, USA \\
  $^{2}$Bangladesh University of Engineering and Technology, Dhaka, Bangladesh \\
  $^{3}$College of William \& Mary, Williamsburg, VA, USA \\
  \texttt{nafeem@rpi.edu} \quad
  \texttt{mjiang04@wm.edu} \quad
  \texttt{hchen23@wm.edu} \quad
  \texttt{yzhang105@wm.edu}
}

\begin{document}
\maketitle
\begin{abstract}
Large language models (LLMs) excel at factual recall yet still propagate stale or incorrect knowledge.
In‑context knowledge editing offers a gradient‑free remedy suitable for black‑box APIs, but current editors rely on static demonstration sets chosen by surface‑level similarity, leading to two persistent obstacles:
(\emph{i}) a quantity–quality trade‑off, and (\emph{ii}) lack of adaptivity to task difficulty.
We address these issues by dynamically selecting \emph{supporting demonstrations} according to their \emph{utility} for the edit.
We propose \textbf{D}ynamic \textbf{R}etriever for \textbf{I}n-Context \textbf{K}nowledge \textbf{E}diting (\textbf{DR-IKE}), a lightweight framework that
(1) trains a BERT retriever with REINFORCE to rank demonstrations by editing the reward, and
(2) employs a \emph{learnable threshold} to prune low‑value examples, shortening the prompt when the edit is easy and expanding it when the task is hard.
DR‑IKE performs editing without modifying model weights, relying solely on forward passes for compatibility with black‑box LLMs. On the \textsc{CounterFact} benchmark, it improves edit success by up to 17.1\%, reduces latency by 41.6\%, and preserves accuracy on unrelated queries, demonstrating scalable and adaptive knowledge editing.
\end{abstract}

\theoremstyle{definition}          
\newtheorem{definition}{Definition}[section]

\section{Introduction}

Large Language Models (LLMs) have demonstrated remarkable capabilities in memorizing vast amounts of knowledge and applying it across various applications, such as question answering \cite{Min2024ExploringTI}, dialogue systems \cite{Feng2023TowardsLD}, and code generation \cite{He2024ImprovingOB}. Nevertheless, because these models are trained on a fixed corpus \cite{Touvron2023Llama2O, Brown2020LanguageMA}, their knowledge often lags behind real‐world developments—either due to erroneous data in the training set \cite{cao2021knowledgeable} or the inherent delay between data collection and model deployment \cite{dhingra2022time,gu2024modeleditingharmsgeneral}. For instance, during a conversation a user might ask, ``\textit{What is the newest iPhone right now?}’’, and a model trained before 2024 could confidently reply, ``\textit{The iPhone 15 is the latest model,}’’ even though Apple introduced the iPhone 16 in September 2024. Such mismatches propagate to end tasks that depend on up-to-date factual grounding, including temporal reasoning \cite{chenghaozhu-etal-2025-llm}, fact verification \cite{Mousavi2024DyKnowDV}, and personalized recommendation \cite{Bao2024RealTimePF}. Retraining a full-scale LLM with fresh data would in principle close this gap, but the computational and financial costs are prohibitive \cite{DeepSeekAI2024DeepSeekV3TR}; as a result, methods for \textbf{Knowledge Editing}—which aim to surgically modify the original model in order to incorporate new or corrected facts \cite{mitchell2021fast}—have emerged as an attractive alternative due to their efficiency and targeted updates.

However, existing practical knowledge editing methods still face three major obstacles. (1)~\emph{Computation cost}. Gradient-based editors such as ROME \cite{Meng2022LocatingAE}, MEND \cite{mitchell2021fast}, MEMIT \cite{meng2023memit}, and SERAC \cite{DeCao2021EditingFK} first calculate the gradient with respect to the parameters that most influence the prediction, and then apply a low-rank update to the implicated weight matrices, so that the model produces the desired answer while leaving other behaviours unchanged.  
Although one edit on GPT-J-6B takes only seconds \cite{meng2023memit}, buffering intermediate activations for T5-11B already consumes $\sim$60 GB of GPU memory \cite{mitchell2021fast}, and both memory and compute scale linearly with model size and the number of edits, rendering thousands of real-time updates on 70-B models infeasible for most labs or start-ups.
Moreover, these algorithms assume full read–write access to the weights, an assumption violated by the growing number of commercial LLM APIs. Consequently, \emph{in-context knowledge editing} (IKE)~\cite{zheng2023ike} has been proposed: it injects the corrected facts as demonstrations in the prompt, thereby achieving competitive success rates without any parameter updates or gradient computations. 
Because most production LLMs are exposed only through black-box APIs, we confine our study to this realistic black-box-only scenario and ask how far a purely in-context approach can be pushed.

Even for in-context-based editors \cite{madaan-etal-2022-memory,zheng2023ike,zhong2023mquake}, two further challenges remain. 
(1)~\emph{Trade-off between example quantity and quality}.  
A series of recent in-context-learning studies reveals that accuracy rises sharply when the prompt includes the first few highly relevant examples, but the marginal benefit of each additional example soon saturates and may even reverse once redundancy or noise creeps in \cite{Chen2023HowMD,Agarwal2024ManyShotIL,zhang2025more,Liu2023LostIT}. In knowledge editing, this fragility is amplified: irrelevant, conflicting, or poorly ordered supporting facts not only fail to help but also propagate errors to otherwise unrelated predictions \cite{yu2024dynamic}, producing ``ripple’’ side-effects \cite{cohen2024evaluating}.
Consequently, practical editors must retrieve and present a minimal set of maximally informative, non-overlapping demonstrations rather than indiscriminately lengthening the prompt.

(2)~\emph{Adaptivity to task difficulty}. Not all edits are equally tractable: recent work shows that success rates vary by knowledge type. Edits involving abstract categories \cite{Ge2024HowWC}, commonsense reasoning \cite{Wu2024AKEWAK}, temporal references \cite{Ge2024TimeSK}, reasoning-heavy logic \cite{Hua2024PropagationAP}, or popular entities \cite{cohen2024evaluating} are consistently more error-prone. In addition, editing multiple related facts increases the risk of contradiction or unintended side-effects \cite{Li2023UnveilingTP}. These findings highlight the need for adjust the editing methods based on the structure and complexity of each task. 

To address the above challenges, we propose \textbf{D}ynamic \textbf{R}etriever for \textbf{I}n-Context \textbf{K}nowledge \textbf{E}diting (\textbf{DR-IKE}), an adaptive example retrieval framework optimized via policy gradients. Our method augments a frozen LLM with a trainable BERT-based retriever that selects and ranks auxiliary factual examples to best facilitate accurate edits. Rather than relying on heuristic retrieval or static prompts, the retriever learns to balance informativeness and factual consistency through interaction: it is rewarded for example selections that lead the LLM to produce the correct, updated output. This approach mitigates the cost and rigidity of gradient-based editors, while also adapting to the difficulty of individual edits through a learnable thresholding mechanism that filters harmful or redundant context. In doing so, our framework offers a scalable, model-agnostic alternative for real-time factual updates, especially in settings with limited access to model internals.

Our main contributions are as follows:

\begin{itemize}
    \item We design a BERT-based retriever trained with policy gradients that selects and ranks auxiliary facts \emph{without touching model weights}. This light-weight, inference-time solution directly avoids the memory and compute overheads for gradient-based editors, making it viable in commercial API settings.

    \item By learning to surface only the most informative, non-overlapping examples, our method curbs prompt length and minimises redundancy, resolving the \emph{quantity–quality trade-off} identified in in-context knowledge editing.

    \item We introduce a dynamic threshold that tightens or relaxes retrieval based on the predicted hardness of each edit, furnishing extra support for abstract, temporal, or popular-entity updates while avoiding over-prompting on easier cases—thereby addressing the need for \emph{adaptivity to task difficulty}.

    \item On the \textsc{CounterFact} benchmark, DR‑IKE outperforms earlier in‑context editors:  
      it trims prompt length, cutting per‑epoch latency by 41.6 \% ;  
      raises Edit‑Success Rate by up to 17.1 \% with fewer tokens;  
      and, thanks to its learnable budget controller, delivers the largest gains on paraphrased and other difficult edits while preserving unrelated knowledge, demonstrating true \emph{adaptivity to task difficulty}.
\end{itemize}

\section{Related Work}

\subsection{In‑context Learning for Knowledge Editing}
Early work on knowledge editing relied on \emph{gradient‑based} parameter updates.  
\citet{dai-etal-2022-knowledge} modify FFN key–value pairs in \textsc{Knowledge Neuron};  
\citet{meng2022locating} apply constrained least squares to the FFN matrix in ROME;  
and \citet{mitchell2021fast} learn low‑rank updates in MEND.  
Although effective, these methods are computationally heavy, can harm unrelated behaviour \cite{gu2024modeleditingharmsgeneral}, and are infeasible for commercial black‑box LLMs \cite{zheng2023ike}.  
This limitation has sparked interest in \emph{in‑context knowledge editing}, which injects the corrected facts as demonstrations in the prompt. Simple prompt engineering—e.g., prefixing with “\textit{Imagine that …}” \cite{cohen2024evaluating}—or chain‑of‑thought prompting in EditCoT \cite{Wang2024KnowledgeET} can inject new facts without touching weights.  
Retrieval‑augmented loops exploit past errors and user feedback: SERAC reroutes matched inputs to a counterfactual model \cite{madaan-etal-2022-memory,mitchell2022memory}, and multi‑hop question formulations improve fidelity \cite{gu2023pokemqa,zhong2023mquake}.  
Demonstration‑diversity strategies—first shown by \citet{si2023prompting} and formalised in IKE via $k$‑NN retrieval of \textsc{Copy}, \textsc{Update}, and \textsc{Retain} examples \cite{zheng2023ike}—boost success rates further.  
Yet existing retrieval pipelines rely solely on embedding similarity, leaving room for \emph{utility-aware} selection and pruning. 
There remains a need for dynamic methods that operate entirely under frozen weights, avoid backpropagation, and minimize both training and inference overhead.

\subsection{Retrieval for In‑context Learning}
Retrieval for ICL traditionally uses static methods such as BM25~\cite{Robertson2009OkapiBM25} or dense embeddings~\cite{Liu2021DenseRetrieval}, which lack task‑aware filtering.  
Fine‑tuned retrievers \citep{Lu2022PromptPG,Li2023UDR,Wang2023LLMR} adapt to specific domains but still return a fixed number of demonstrations.  
\citet{yu2024dynamic} model retrieval as a Markov Decision Process with a learnable threshold for long‑tail QA. While their bag-of-words preselection suffices for that setting, it lacks the context sensitivity needed for knowledge editing. We adapt this framework by using embeddings-based preselection for richer semantics and editing-specific rewards to encourage efficacy, specificity, and stability.

\section{Problem Statement}

\begin{figure}[h]
  \centering
  \includegraphics[width=0.5\textwidth]{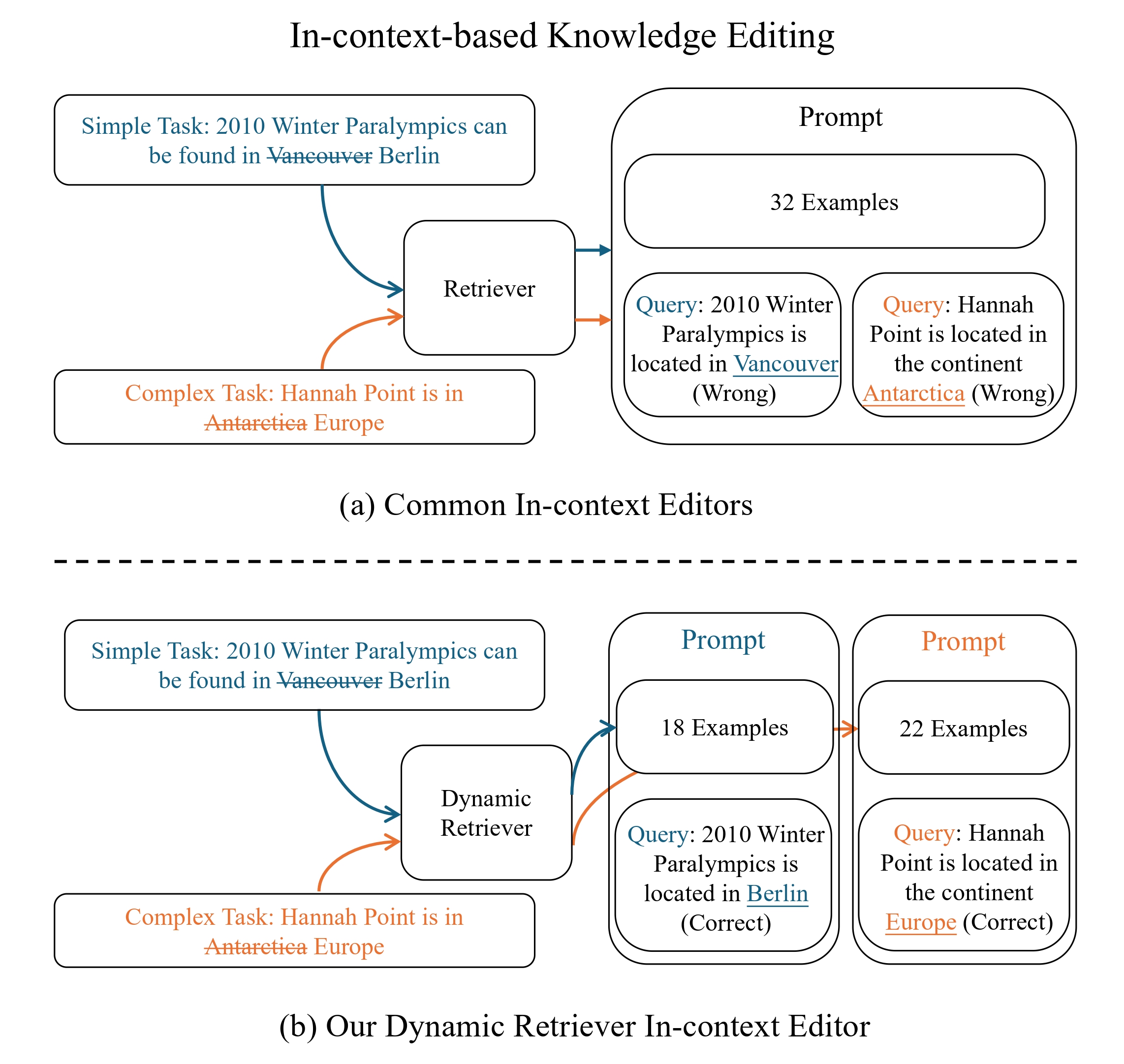}  
\caption{Motivating example for adaptive in‑context editing. 
(a)~A \emph{fixed‑shot} editor appends the same large set of demonstrations to every query; the oversized prompt diffuses model attention, hindering both a straightforward location swap (\textit{Winter Paralympics}: Vancouver~$\rightarrow$~Berlin) and a more concept‑heavy update (\textit{Hannah Point}: Antarctica~$\rightarrow$~Europe), whose corpus evidence is dominated by Antarctic descriptions. 
(b)~Our \emph{dynamic} editor adjusts prompt length to task complexity, retaining a minimal context for the simple edit while retrieving additional, targeted facts for the harder one.}

  \label{fig:fig1}
\end{figure}

\begin{definition}[Knowledge Editing]
Let $\mathcal{M}$ be a frozen large‑language model and let $\mathcal{K}_{c}$ denote a single factual triple (or small set of triples) stored in the model’s parametric memory.  
Knowledge editing seeks a post‑edited model $\mathcal{M}' = f(\mathcal{M},\mathcal{K}_{c}\!\rightarrow\!\mathcal{K}_{c}')$ such that

\begin{enumerate}
    \item for any query $q$ whose answer depends solely on $\mathcal{K}_{c}$, the response of $\mathcal{M}'$ is consistent with the revised fact $\mathcal{K}_{c}'$, and
    \item for any query that depends on unrelated knowledge $\mathcal{K}_{s}$ ($\mathcal{K}_{s}\cap\mathcal{K}_{c}=\varnothing$), the behavior of $\mathcal{M}'$ matches that of the original model~$\mathcal{M}$.
\end{enumerate}
\end{definition}

\begin{definition}[In‑context Knowledge Editing]
A language model may be modified either by changing its parameters or by altering the input prompt $P$.  
In‑context Knowledge Editing keeps the parameters of $\mathcal{M}$ frozen and employs a retriever $R$ to construct an augmented In-Context Learning (ICL) prompt
\(
P^{\star}=P\!+\!\langle d_{1},\dots,d_{m}\rangle,
\)
where $d_{j}$ are natural‑language \emph{demonstrations}.  
Conventional editors choose a \emph{fixed} number \(m\!\ge\!0\) of demonstrations~\cite{zheng2023ike,cohen2024evaluating}.  
We generalize this paradigm by learning a threshold $\sigma$ that
allows the retriever to select a \emph{variable} number
\(
k_\sigma(q)=\bigl|\{j \mid \pi_j\!\ge\!\sigma\}\bigr|
\)
of demonstrations, where \(\pi_j\) is the retriever score of candidate \(j\);
see Fig.~\ref{fig:fig1}(b) for an illustration.

\end{definition}

\begin{definition}[Demonstration Categories \cite{zheng2023ike}]
Each retrieved example is assigned one of three functional roles:
\end{definition}

\begin{itemize}
  \item \textbf{Copy}—explicitly restates the target fact to reinforce $\mathcal{K}_{c}'$.\\
        Example: “\textit{The official language of Brazil is} $\rightarrow$ Spanish.”
  \item \textbf{Update}—paraphrases the query before introducing the new fact.\\
        Example: “\textit{Brazil’s national language has been changed to} $\rightarrow$ Spanish.”
  \item \textbf{Retain}—references a related context that \emph{should not} change.\\
        Example: “\textit{The official language of Canada is} $\rightarrow$ English.”
\end{itemize}

While \textsc{Retain} examples increase specificity, an excess of low‑utility \textsc{Retains} can bloat the prompt and push the LLM to fall back on outdated parametric memory.  
Therefore, our method dynamically selects only a subset of the \textsc{Retain} pool, while all three categories remain available for prompt construction.

\begin{figure*}[ht!]
  \centering
  \includegraphics[width=0.8\textwidth]{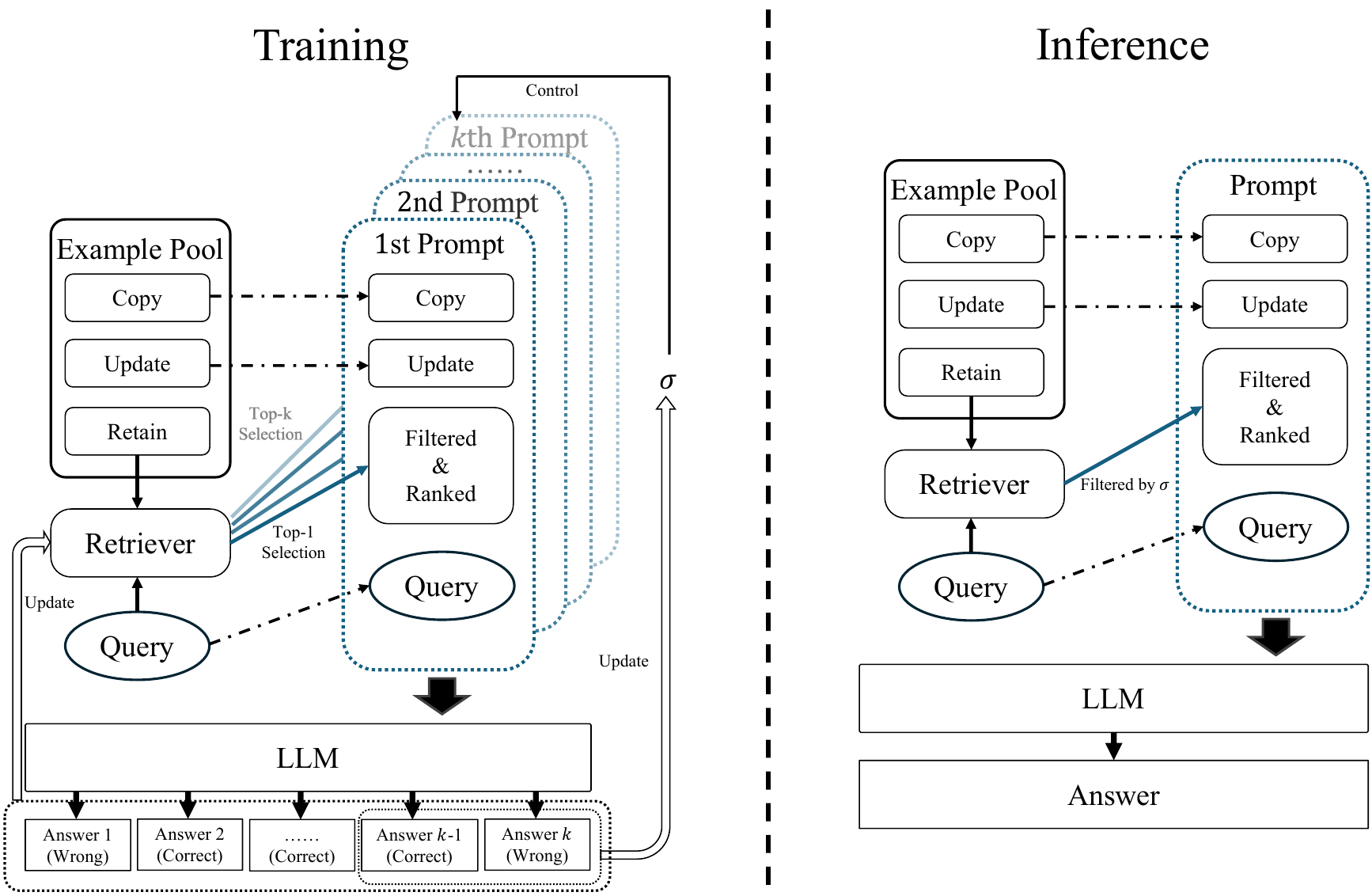}

\caption{Overall architecture of the proposed framework.  
\textbf{Training stage (left).}  
Given an edit query, the retriever scores a pool of retained candidate demonstrations.  
A learnable threshold~\(\sigma\) converts these scores into an integer \(k\), the number of prompt variants to generate.  
The frozen LLM processes each variant; its outputs yield the reinforcement signal used to update the retriever via policy gradients.  
Whenever the aggregated outcome flips between success and failure, \(\sigma\) is adjusted to capture the revised difficulty estimate.  
\textbf{Inference stage (right).}  
At test time, the learned \(\sigma\) truncates the ranked list once, yielding a single concise prompt to the frozen LLM.}

  \label{fig:main}
\end{figure*}
  
\section{Method}

We now give a brief tour of \textbf{DR‑IKE} (see Fig.~\ref{fig:main} for a schematic). During training, a lightweight BERT retriever learns, via REINFORCE \cite{Williams1992}, to rank \textsc{Retain} candidates, while a learnable threshold~$\sigma$ simultaneously adjusts the number of examples admitted to the prompt.  
At inference time the same threshold truncates the ranked list once, producing a single compact prompt that is sent to the black‑box LLM.

The rest of this section is organized as follows.  
Section \ref{subsec:mdp} formalizes the retrieval process as a Markov Decision Process, specifying the state, action, policy, and reward.  
Section \ref{subsec:training} details the policy‑gradient training protocol.  
Finally, Section \ref{subsec:budget} introduces the dynamic budget controller that adapts~$\sigma$ to prevent prompt bloat.

\subsection{Markov‑Decision‑Process Formulation \label{subsec:mdp}}

We cast \textsc{Retain}-example selection as an (Markov Decision Process) MDP  
\(\langle\mathcal{S},\mathcal{A},T,R\rangle\) so that the retriever can optimize for \emph{utility} rather than raw similarity.

\paragraph{State.}
At step \(t\) the state is  
\(s_t=(x,\mathcal{R}_t)\), where \(x\) is the query and  
\(\mathcal{R}_t=\{e_{1},\dots ,e_{t}\}\subseteq\mathcal{C}\) is the ordered set of \textsc{Retains} chosen so far.  
The candidate pool  
\(\mathcal{C}=\{e^{(1)},\dots ,e^{(n)}\}\) is produced off‑line by KNN retrieval over sentence embeddings.

\paragraph{Action.}
The action space is \(\mathcal{A}=\mathcal{C}\cup\{\textit{stop}\}\).  
Choosing \(a_t=e^{(i)}\) appends that example to the growing prompt;  
choosing \textit{stop} finalizes the prompt and terminates the episode.

\paragraph{State Transition.}
If \(a_t\neq\textit{stop}\) the transition is deterministic:
\begin{equation}
    s_{t+1}=(x,\mathcal{R}_t\cup\{a_t\}).
\end{equation}
No successor state is generated after \textit{stop}.

\paragraph{Reward.}
After each action $a_{t}$, the current prompt is fed to the frozen LLM and a step‑wise reward is issued:
\begin{equation}
r_{t+1} \;=\; 2 \times \mathbf{1}\!\bigl[\hat y_{t+1} = y_{\text{new}}\bigr]-1,
\label{eq:reward}
\end{equation}
where \(\hat y_{t+1}\) is the LLM’s answer and \(\mathbf{1}[\cdot]\) is the indicator. This step-wise feedback guides the retriever in estimating the marginal impact of each added example on editing success.

\subsection{Policy‑gradient Training of the Retriever}\label{subsec:training}

Fig.~\ref{fig:main}\,(left) outlines the learning loop.  At each epoch we optimize the retriever parameters while the LLM remains frozen.

\paragraph{Demonstration selection.}
For each edit instance \((x, y_{\text{new}})\), we use a pretrained 20M-parameter Sentence-Transformer~\cite{reimers-2020-sbert} to retrieve fixed \textsc{Copy} and \textsc{Update} demonstrations, and to preselect a \emph{candidate pool} for \textsc{Retain} examples.

\paragraph{Policy and action sampling.}
The retriever is a frozen 4‑layer BERT encoder (29 M parameters) \cite{turc2019} followed by a trainable linear head
\(S_{\theta}\) whose parameters are denoted by \(\theta\).
Each \(e^{(i)}\in\mathcal{C}\) receives a score
\(z_i=S_{\theta}(x,e^{(i)})\).
A softmax yields the categorical policy: 
\begin{equation}
    \pi_{\theta}(a_t=e^{(i)}\mid s)=\frac{\exp(z_i)}{\sum_{j=1}^{n}\exp(z_j)}.
\end{equation}
At step \(t\) an index \(a_t\) is sampled from \(\pi_{\theta}\) and the corresponding retain
\(e_t\) is appended to the prompt \emph{iff}
\(\pi_{\theta}(a_t\mid s_t)>\sigma\); otherwise the threshold mechanism (Section~\ref{subsec:budget}) emits the \textit{stop} action and the episode terminates.

\paragraph{Policy Update.}
After each action the current prompt is queried and a binary reward \(r_t\in\{+1,-1\}\) is observed Eq. \ref{eq:reward}.  
The REINFORCE \cite{Williams1992} loss is

\begin{equation}
\mathcal{L}(\theta)= -\sum_{t=1}^{T} r_t\,\log \pi_{\theta}(a_t\mid s_t),
\end{equation}

\noindent
where \(T\) is the (variable) episode length.  
Because gradients flow only through the linear head, updates are computationally light while still teaching the retriever \emph{which} retains to keep and \emph{when} to stop.  
A complete, line‑by‑line pseudocode listing appears in Appendix~\ref{appendix:training_protocol}.


\begin{table*}[h]
  \centering
  \begin{tabular}{|l|c|c|c|c|c|c|c|c|}
    \hline
    \textbf{Editing Method} & \textbf{Extra Params} & \textbf{S ↑} & \textbf{ESR ↑} & \textbf{PC ↑} & \textbf{RR ↑} & \textbf{ESM ↑} & \textbf{GSM ↑} & \makecell{\textbf{Inference}\\\textbf{Time(s) ↓ }} \\ 
    \hline
    \multicolumn{9}{|l|}{\textbf{Llama-3.1-8B-Instruct}} \\ 
    \hline
    FactPrompt             & 0   & 0.434 & 0.61 & 0.34 & 0.43 & 0.21 & -0.05 & \textbf{1.99} \\ 
    \hline
    EditCoT                & 0   & 0.431 & 0.70 & 0.33 & 0.40 & 0.24 & -0.06 & \underline{2.02} \\ 
    \hline
    IKE                    & 20M & \underline{0.727} & \underline{0.76} & \underline{0.67} & \textbf{0.76} & \underline{0.43} & \underline{0.39} & 6.52 \\ 
    \hline
    \textbf{DR-IKE (Ours)} & 49M & \textbf{0.775} & \textbf{0.89} & \textbf{0.81} & \underline{0.66} & \textbf{0.69} & \textbf{0.49} & 3.81 \\ 
    \hline
    \multicolumn{9}{|l|}{\textbf{Qwen 2.5-7B}} \\ 
    \hline
    FactPrompt             & 0   & 0.335 & 0.48 & 0.26 & 0.33 & 0.54 & 0.45 & \underline{2.42} \\ 
    \hline
    EditCoT                & 0   & 0.424 & 0.53 & 0.43 & 0.35 & -0.09 & -0.15 & \textbf{2.25} \\ 
    \hline
    IKE                    & 20M & \underline{0.738} & \underline{0.75} & \underline{0.69} & \textbf{0.78} & \underline{0.64} & \underline{0.57} & 6.75 \\ 
    \hline
    \textbf{DR-IKE (Ours)} & 49M & \textbf{0.779} & \textbf{0.89} & \textbf{0.77} & \underline{0.70} & \textbf{0.83} & \textbf{0.74} & 4.21 \\ 
    \hline
  \end{tabular}
  \caption{Editing performance across methods for two base models, including extra parameter counts, harmonic mean \(S\), edit/generalization success margins (ESM/GSM), and measured inference time per iteration. Best scores are in \textbf{bold}, second-best are \underline{underlined}.}
  \label{tab:editing_methods}
\end{table*}

\subsection{Dynamic Budget Controller}\label{subsec:budget}
\vspace{-4pt}
To curb prompt bloat we endow the retriever with a \emph{learnable threshold}~$\sigma$ called budget controller that caps the number of \textsc{Retain} examples.  
At the start of training we set $\sigma=0$, guaranteeing that at least one \textsc{Retain} can be selected.  
During both training and inference a candidate $e^{(i)}$ is kept only if its policy probability satisfies $\pi_{\theta}(a_t=e^{(i)}\mid s)>\sigma$; candidates are considered in descending order of probability.
\paragraph{Adaptive update.}
While constructing the prompt we monitor the binary reward $r\!\in\!\{+1,-1\}$ after each newly added example.  
If appending the $(j{+}1)$‑th \textsc{Retain} turns a previously correct answer into an incorrect one, we tighten the budget by raising~$\sigma$ to the largest probability among the \emph{remaining} candidates:
\begin{equation}
\sigma \;\leftarrow\;
\max\!\Bigl(\,\sigma,\;
\max_{\,i>j}\,\pi_{\theta}(a_t=e^{(i)}\mid s)\Bigr).
\end{equation}
\noindent
Thus the bar for inclusion becomes progressively higher on difficult edits, whereas easy edits naturally terminate after a few high‑utility examples.  
This single scalar threshold allows the system to learn \emph{both} which \textsc{Retain} matter and \emph{how many} are worth keeping, achieving compact yet effective prompts.

\section{Experiments}
\subsection{Experimental Setup}

\paragraph{Dataset} 

We use the \textsc{CounterFact} benchmark~\citep{meng2022locating}, a widely adopted evaluation suite for factual knowledge editing in language models. It comprises 21,919 factual records. We follow~\citet{zheng2023ike} in using the first 2,000 records for the editable sample pool and the remainder for constructing ICL demonstrations.

\paragraph{Language Model}
We use Meta-Llama-3.1-8B-Instruct and Meta-Llama-3.2-3B-Instruct~\cite{grattafiori2024llama3}, Mistral-7B-Instruct-v0.2~\cite{fioravanti2023mistral}, and both sizes of Qwen 2.5 (7B and 1.5B)~\cite{zhang2024qwen25} via HuggingFace’s pipeline API. The model parameters are not updated at any point. 

\paragraph{Training Configuration} 
We train on 300 randomly selected samples and evaluate on 100, using the Adam optimizer \cite{DBLP:journals/corr/KingmaB14} with a learning rate of \(1 \times 10^{-4}\) for 5 epochs. Each training episode corresponds to a query, with batch size 1 for step-wise updates guided by policy-based optimization.

\paragraph{Compute Environment} 
All training was conducted on Google Colab using an NVIDIA L4 GPU (24 GB VRAM), optimized for inference and lightweight training. We implement our pipeline in PyTorch and HuggingFace Transformers v4.39.3.

\paragraph{Baselines} 
We compare our method with three representative in‐context editing strategies: \textbf{FactPrompt} \cite{cohen2024evaluating}, which steers the model by prepending a narrative prefix (“Imagine that…”); \textbf{EditCoT} \cite{Wang2024KnowledgeET}, which applies chain‐of‐thought prompting to guide step‐by‐step reasoning before editing; and \textbf{IKE} \cite{zheng2023ike}, which iteratively selects \textsc{Copy}, \textsc{Update}, and \textsc{Retain} examples to inject new facts while preserving existing knowledge.

\paragraph{Evaluation Metrics} 
We evaluate editing performance using standard metrics introduced by \citet{zheng2023ike}. \textit{\textbf{Edit Success Rate (ESR)}} measures how often the LLM outputs the correct edited object (\texttt{target\_new}); \textit{\textbf{Retention Rate (RR)}} measures whether the original object (\texttt{target\_true}) is preserved on unrelated neighborhood prompts; and \textit{\textbf{Paraphrase Consistency (PC)}} checks agreement on paraphrased in-scope prompts. We report their harmonic mean as \textit{\textbf{Score (S)}} following \citet{meng2022locating}. We also include \textit{\textbf{Edit Success Magnitude (ESM)}} and \textit{\textbf{Generalization Success Magnitude (GSM)}}, which quantify log-probability shifts between \texttt{target\_new} and \texttt{target\_true} on edited and paraphrased prompts, respectively. ESR is tracked epoch-wise to monitor retriever learning.

\subsection{Main Results}
Tab.~\ref{tab:editing_methods} shows the performance of knowledge editing in different methods using two different LLMs, Llama 3.1 Instruct-8b and Qwen 2.5-7b. 

As expected, all methods yield comparable ESR scores, but diverge sharply on PC and RR. FactPrompt and EditCoT, for example, perform reasonably on ESR (0.61 and 0.70) but falter on PC and RR due to their limited prompt design. Without \textsc{Update} or \textsc{Retain} examples, the LLM lacks guidance on when to generalize or preserve original knowledge. The IKE method improves on this by incorporating \textsc{Copy}, \textsc{Update}, and \textsc{Retain} examples, enabling more structured control over generalization and specificity. However, its inclusion of all \textsc{Retains} can lead to noisy prompts that hinder effective editing. 

Our method addresses this by using a retriever trained via policy-based optimization to rank \textsc{Retains} and a budget controller to keep only the most helpful ones. As a result, it shows noticeable improvement over ESR and PC while remaining competitive on RR across both LLMs. We also observe consistent gains in \textit{ESM} and \textit{GSM}—which measure the model’s confidence in the edited and generalized responses, respectively—indicating that DR-IKE not only produces correct outputs but does so with higher certainty. 

Additionally, comparing the last column, our method matches or outperforms IKE across all but one metric while achieving lower inference time, thanks to the budget controller’s adaptive prompt shortening.

\setlength{\tabcolsep}{12pt}
\begin{table}[htb]
  \centering
  \begin{tabular}{|l|c|c|c|}
    \hline
    \textbf{Method}       & \textbf{ESR ↑} & \textbf{PC ↑} & \textbf{RR ↑} \\ 
    \hline
    \multicolumn{4}{|l|}{\textbf{Llama-3.1-8B-Instruct}} \\ 
    \hline
    IKE-All               & 0.76           & 0.67          &\underline{0.76}          \\ 
    \hline
    Rank-All              & 0.81           & 0.69          & \underline{0.76}         \\ 
    \hline
    Rank-50\%             & \underline{0.86}           & \underline{0.71}          & \textbf{0.78}          \\ 
    \hline
    \textbf{DR-IKE}       & \textbf{0.89}           & \textbf{0.81}          & 0.66          \\ 
    \hline
    \multicolumn{4}{|l|}{\textbf{Qwen-2.5-7B-Instruct}} \\ 
    \hline
    IKE-All               & 0.75           & 0.69          & \underline{0.78}          \\ 
    \hline
    Rank-All              & 0.84           & 0.67          & \underline{0.78}          \\ 
    \hline
    Rank-50\%             & \underline{0.84}           & \underline{0.76}          & \textbf{0.79}          \\ 
    \hline
    \textbf{DR-IKE}       &\textbf{0.89}           & \textbf{0.77}          & 0.70          \\ 
    \hline
  \end{tabular}
  \caption{Ablation of \textsc{Retain}-example selection: IKE-All (baseline), static ranking (Rank-All, Rank-50\%), and our dynamic controller (DR-IKE).}
  \label{tab:retain_selection_ablation}
\end{table}

\begin{figure*}[h]
  \centering
  \includegraphics[width=\textwidth]{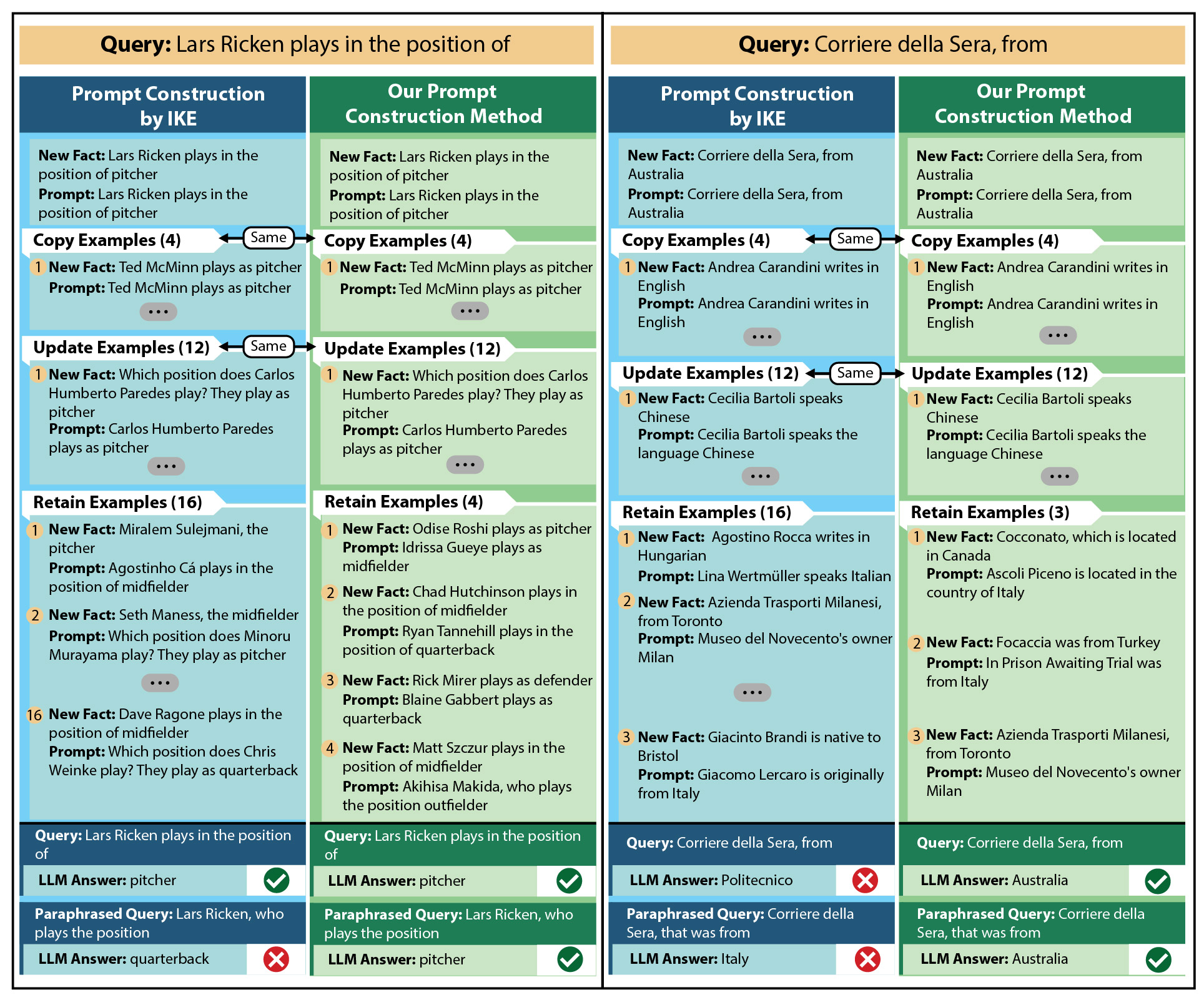}
  \caption{Case study illustrating prompt construction and example selection for challenging queries under IKE versus our method.}
  \label{fig:case_study_examples}
\end{figure*}
\subsection{Ablation Study of the Retain‐Example Budget Controller}
Tab.~\ref{tab:retain_selection_ablation} contrasts four \textsc{Retain}‐selection strategies. \textbf{IKE-All} uses all examples without filtering. \textbf{Rank-All} ranks and keeps all, i.e. no pruning of redundant \textsc{Retain} examples. \textbf{Rank-50\%} ranks and keeps only the top half, adding a static budget. \textbf{DR-IKE} uses our dynamic controller to prune low-utility examples on the fly, yielding the best balance of efficiency and efficacy.

\textbf{Rank-All} yields performance comparable to \textbf{IKE-All}, indicating that ordering alone is insufficient to improve editing outcomes. Introducing a static budget with \textbf{Rank-50\%} which ranks and keeps only the top half of \textsc{Retain} examples enhances ESR, PC, and RR, confirming that an overabundance of redundant demonstrations can degrade efficacy, generalization, and specificity. Finally, \textbf{DR-IKE} uses our dynamic controller to prune low-utility examples on the fly, further boosting ESR and PC by adaptively selecting the most informative \textsc{Retains}; the modest drop in RR suggests that overly stringent pruning may occasionally discard useful context. Overall, dynamic example budgeting via DR-IKE proves essential for achieving the optimal balance of efficiency and effectiveness in knowledge editing. For empirical statistics on the number of \textsc{Retain} examples selected per prompt across 100 test instances, refer to Appendix~\ref{appendix:retain_budget}.

\paragraph{Effects on Training Efficiency}
Introducing our dynamic budget controller substantially reduces training time. By pruning lower-value \textsc{Retain}, we shorten prompts and cut LLM inference overhead. For Llama 3.1-8B, per-epoch runtime drops from 2080.1s to 1459.6s (29.8\% saving), and for Qwen 2.5-7B from 1459.6s to 1174.6s (19.5\%). This efficiency gain supports scalable deployment on larger models.  

\subsection{Case Study}
To intuitively show the effectiveness of our proposed method, we show two knowledge editing case studies from the Counterfact dataset in Fig.~\ref{fig:case_study_examples}. The edited fact, \textsc{Copy} and \textsc{Update} examples remain the same for both methods. In IKE (blue columns), \textsc{Retain} examples are fixed, whereas in our method (green columns), they are dynamically selected using the retriever and budget controller. In the first case study, IKE’s prompt construction correctly guides the LLM on the original query but fails when the prompt is paraphrased, whereas our method succeeds on both. This difference arises because our retriever identifies \textsc{Retain} examples whose contextual cues closely align with the factual update, regardless of surface variation, ensuring that even paraphrased queries receive the appropriate support. In the second case study, IKE fails on both the original and paraphrased queries—likely because its \textsc{Retain} examples mix language-based rather than location-based contexts—while our method retrieves only location-based \textsc{Retains}, resulting in correct answers in both cases and confirming the suitability of our example selection.
\setlength{\tabcolsep}{6pt}
\begin{table}[ht]
  \centering
  \begin{tabular}{|l|c|c|c|}
    \hline
    \textbf{Model (Parameters)}      & \textbf{ESR ↑} & \textbf{PC ↑} & \textbf{RR ↑} \\ 
    \hline
    Llama 3.1 (8B)          & \textbf{0.89}           & \textbf{0.81}          & \underline{0.66}          \\ 
    \hline
    Llama 3.2  (3B)          & \underline{0.86}           & 0.71          & 0.61          \\ 
    \hline
    Mistral v0.2 (7B)       & 0.53           & 0.42          & 0.34          \\ 
    \hline
    Qwen 2.5 (7B)                    & \textbf{0.89}           & \underline{0.77}          & \textbf{0.70}          \\ 
    \hline
    Qwen 2.5 (1.5B)                  & 0.57           & 0.42          & 0.39          \\ 
    \hline
    SmolLM2 (1.7B)                  & 0.46           & 0.27          & 0.22          \\ 
    \hline
  \end{tabular}
  \caption{Performance comparison of various LLMs.}
  \label{tab:llm_performance}
\end{table}
\subsection{Base Model Performance}
Editing performance often hinges on a model’s capacity to integrate new information while preserving existing knowledge. Tab.~\ref{tab:llm_performance} shows that smaller LLMs, e.g.\ \textbf{Llama 3.2 (3B)}, can match larger counterparts in ESR but suffer noticeable drops in PC and RR, a pattern echoed by the tiny \textbf{Qwen 1.5B} and \textbf{SmolLM2 1.7B}. This underscores the role of scale in furnishing the nuanced representations needed for precise, stable edits. \textbf{Mistral v0.2 (7B)} underperforms across all metrics, likely because its limited $8$k context window restricts the in-context learning needed for effective editing \citep{xu2024retrieval}. Moreover, instruct‐fine‐tuned models like \textbf{Llama 3.1 (8B)} and \textbf{Qwen 2.5 (7B)} demonstrate particular resilience to paraphrased prompts, more reliably retaining injected facts under syntactic variation than their smaller or non‐instruct counterparts.

\subsection{DR-IKE under Black-Box API Settings}
We additionally tested \textsc{DR-IKE} under black-box conditions using the 
\href{https://www.kaggle.com/models/google/gemini-2.0-flash-api/Api/gemini-2.0-flash/1}{Gemini-2.0-Flash API on Kaggle}. 
As shown in Table~\ref{tab:blackbox_eval}, \textsc{DR-IKE} achieved 
substantial gains in Edit Success Rate (ESR) and Paraphrase Consistency (PC) compared to the in-context editing baseline (\textsc{IKE}). 
The improvements are not only considerable in absolute terms, but they also surpass the margins we observed on earlier experiments with smaller open-weight models such as \textsc{LLaMA}-3.1-8B or Mistral-7B. 
This suggests that larger LLMs may benefit even more from the dynamic retrieval strategy employed by \textsc{DR-IKE}. 
At the same time, we note a decrease in Retain Rate (RR), indicating a trade-off between successful edits, generalization, and preservation of unrelated knowledge. 
Future work could explore techniques to further stabilize RR without compromising the strong gains in ESR and PC.
\setlength{\tabcolsep}{12pt}
\begin{table}[htb]
  \centering
  \begin{tabular}{|l|c|c|c|}
    \hline
    \textbf{Method}       & \textbf{ESR ↑} & \textbf{PC ↑} & \textbf{RR ↑} \\ 
    \hline
    IKE                   & 0.69           & 0.52          & \textbf{0.57} \\ 
    \hline
    \textbf{DR-IKE}       & \textbf{0.91}  & \textbf{0.83} & 0.46          \\ 
    \hline
  \end{tabular}
  \caption{Black-box evaluation of \textsc{DR-IKE} against the in-context editing baseline (\textsc{IKE}) using the \href{https://www.kaggle.com/models/google/gemini-2.0-flash-api/Api/gemini-2.0-flash/1}{Gemini-2.0-Flash API on Kaggle}.}
  \label{tab:blackbox_eval}
\end{table}

\subsection{Extended Evaluation on Benchmark Datasets}

We further evaluated \textsc{DR-IKE} on two additional benchmarks: zsRE~\cite{meng2022locating} and WikiDataCounterFact~\cite{zhong2023mquake}. Both datasets include a broader range of edit types and formats. The same experimental setup as in the main paper was used, with the \textsc{LLaMA-3.1-8B} model. Notably, the average similarity to $k$-nearest neighbors is substantially lower in zsRE (0.4042) and WikiDataCF (0.4507) than in CounterFact (0.5695), indicating that these datasets contain more diverse and less redundant examples. This demonstrates that \textsc{DR-IKE} maintains strong performance even in lower-similarity, low-resource settings. 

As shown in Table~\ref{tab:extended_eval}, \textsc{DR-IKE} achieves consistently higher Edit Success Rate (ESR) and Paraphrase Consistency (PC) compared to the in-context editing baseline (\textsc{IKE}). Retain Rate (RR) is also reported alongside ESR and PC, providing a more complete picture of the trade-off between successful edits, generalization, and preservation of unrelated knowledge.

\setlength{\tabcolsep}{12pt}
\begin{table}[htb]
  \centering
  \begin{tabular}{|l|c|c|c|}
    \hline
    \textbf{Method}       & \textbf{ESR ↑} & \textbf{PC ↑} & \textbf{RR ↑} \\ 
    \hline
    \multicolumn{4}{|l|}{\textbf{zsRE}} \\ 
    \hline
    IKE                   & 0.30           & 0.22          & 0.49          \\ 
    \hline
    \textbf{DR-IKE}       & \textbf{0.33}  & \textbf{0.26} & \textbf{0.51} \\ 
    \hline
    \multicolumn{4}{|l|}{\textbf{WikiDataCounterFact}} \\ 
    \hline
    IKE                   & 0.39           & 0.40          & 0.63          \\ 
    \hline
    \textbf{DR-IKE}       & \textbf{0.42}  & \textbf{0.43} & \textbf{0.64} \\ 
    \hline
  \end{tabular}
  \caption{Extended evaluation of \textsc{DR-IKE} on zsRE and WikiDataCounterFact.}
  \label{tab:extended_eval}
\end{table}

\section{Conclusion}
In this work, to improve upon existing demonstration strategies and prompting paradigms for knowledge editing, we propose Dynamic Retriever for In-Context Knowledge Editing (DR-IKE), which uses a pre-trained BERT-based retriever (trained via policy gradient) to rank examples by their editing utility and a budget controller to prune lower-value cases from the prompt. In particular, we examine \textsc{Retain} examples that, while safeguarding auxiliary context, may compromise edit efficacy; our retriever instead prioritizes those that genuinely bolster both fact injection and knowledge preservation, as confirmed by our empirical results. DR-IKE’s combination of learned ranking and adaptive budgeting yields slightly higher edit success and consistency than leading in-context editing methods, while also reducing inference and training time. These results suggest that DR-IKE can scale effectively to very large, black-box LMs. Future work could extend DR-IKE to better handle fact-type variation, low-overlap retrieval settings, and complex edits by incorporating fact-aware retrieval and more flexible retrieval sources.

\section*{Limitations}
There are several limitations of our work. First, our evaluation is constrained by the COUNTERFACT benchmark, which does not categorize facts by type (e.g.\ historical, numerical, geographical, technical). As a result, we cannot assess how editing performance varies across different knowledge domains. Furthermore, DR-IKE relies on the presence of sufficiently similar paraphrases and neighborhood examples within a single dataset. In scenarios where such examples are sparse or absent, retrieval quality and editing efficacy may degrade, limiting the method’s applicability in low-resource or highly specialized domains. Finally, although our budget controller mitigates prompt-length concerns, LLM context windows remain finite. In cases requiring a large number of demonstrations—such as nuanced multi-step edits or extensive domain coverage—dynamic budgeting alone may not suffice to fit all necessary examples within the model’s maximum input length.  

\section*{Acknowledgement}
This work is supported in part by the National Science Foundation (NSF) grant IIS-2451436 and Commonwealth Cyber Initiative grant HC-4Q24-059.

\bibliography{custom}

\appendix
\section{Appendix: Training Procedure for the Dynamic Retriever}
\label{appendix:training_protocol}

\begin{algorithm}[htb!]
\caption{Training Protocol for Dynamic Example Selection}
\label{alg:dynamic_example_selection}
\begin{algorithmic}[1]
\Require
  initial parameters $\theta_0$;\; training set $\mathcal{T}$;\;
  max shots $K$;\; learning rate $\alpha$
\Ensure
  trained retriever $\theta$ and budget threshold $\sigma$
\State $\theta \gets \theta_0$, \quad $\sigma \gets 0$
\For{\textbf{epoch} $=1$ \textbf{to} $N_{\text{epochs}}$}
  \For{\textbf{each} $(x,\,y_{\text{new}})\in\mathcal{T}$}
    \State Fixed \textsc{Copy} / \textsc{Update} demos for $x$
    \State $\mathcal{C}\leftarrow$ \textsc{Retain} candidates 
    \Statex \hspace{5em}(pre-selected via embeddings)
    \State $z \gets S_{\theta}(x,\mathcal{C})$  \Comment scalar scores
    \State $p \gets \mathrm{softmax}(z)$ \Comment policy distribution
    \State $k \gets \min\!\bigl(K,\,|\{i \mid p_i>\sigma\}|\bigr)$
    \If{$k = 0$} \State $k \gets 1$ \EndIf
    \State $\textit{prev}\_r \gets \bot$, \quad $\mathcal{L}\gets 0$
    \For{$j = 1$ \textbf{to} $k$}
      \State $\mathcal{R}_j \gets$ top-$j$ \textsc{Retains} from $\mathcal{C}$ by $p$
      \State Prompt with \textsc{Copy}, \textsc{Update}, $\mathcal{R}_j$
      \State query LLM $\rightarrow \hat{y}$
      \State $r \gets \begin{cases}
              +1,& \hat{y}=y_{\text{new}}\\
              -1,& \text{otherwise}
            \end{cases}$

      \State $c \gets (j>1)\wedge(\textit{prev\_r}=+1)$

      \Statex \hspace{6.5em}$\wedge(r=-1)$
      \If{$c$}
            \State $\sigma \gets \max\!\bigl(\sigma,\;\max_{i>j}p_i\bigr)$
      \EndIf

      \State $\mathcal{L} \mathrel{+}= -\,r\;\log p_j$
      \State \textit{prev}\_r $\gets r$
    \EndFor
    \State $\theta \gets \theta - \alpha\,\nabla_{\theta}\mathcal{L}$ \Comment REINFORCE
  \EndFor
\EndFor
\end{algorithmic}
\end{algorithm}

The overall training protocol for our dynamic retriever is illustrated in Algorithm~\ref{alg:dynamic_example_selection}. At the beginning of each epoch, we initialize the retriever’s parameters \(\theta\) and the budget threshold \(\sigma\). For every training instance \((x, y_{\mathrm{new}})\) in the dataset, we first construct the fixed \textsc{Copy} and \textsc{Update} demonstrations based on semantic similarity using embeddings from a 20M-parameter Sentence Transformer~\cite{reimers-2020-sbert}. These fixed components serve as the foundational part of every in-context prompt. Next, we select a candidate pool \(\mathcal{C}\) of \textsc{Retain} examples—again using the same embedding space—to identify potentially helpful facts related to the input.

Once the \textsc{Retain} pool is constructed, the retriever’s scoring function \(S_\theta\) evaluates each candidate, producing scalar relevance scores which are normalized via softmax to obtain a probability distribution \(p\) over the pool. The retriever then determines how many examples to select, based on the current value of \(\sigma\), by choosing the top candidates whose probabilities exceed the threshold. If none of the probabilities surpass \(\sigma\), we ensure that at least one \textsc{Retain} is selected to prevent empty prompt construction. This mechanism enables the retriever to modulate prompt length dynamically, balancing retrieval confidence with context limitations.

We then proceed to construct the full in-context prompt incrementally by adding one \textsc{Retain} example at a time (top-\(j\) based on \(p\)). After each addition, we query the language model and check if its output matches the desired edited response \(y_{\mathrm{new}}\). A binary reward \(r \in \{+1, -1\}\) is assigned depending on whether the answer is correct. If adding a particular \textsc{Retain} causes a correct prediction to flip to incorrect, we treat it as a degradation in prompt quality and increase the threshold \(\sigma\) to exclude lower-probability examples in future steps. This allows the model to prune less useful \textsc{Retains} early and adapt the prompt construction strategy over time.

Throughout the episode, we accumulate REINFORCE-style loss \cite{Williams1992} using the reward signal and the log-probability of each selected \textsc{Retain}. The resulting gradient is used to update only the retriever’s scoring head via policy gradient. By repeating this process across training epochs, the retriever learns not just which \textsc{Retain} examples are most helpful for factual editing, but also how to dynamically adjust its selection budget to optimize both model performance and context efficiency.

\section{Appendix: Retain Budget Distribution Across Models}
\label{appendix:retain_budget}

\begin{figure}[h]
  \centering
  \includegraphics[width=0.8\linewidth]{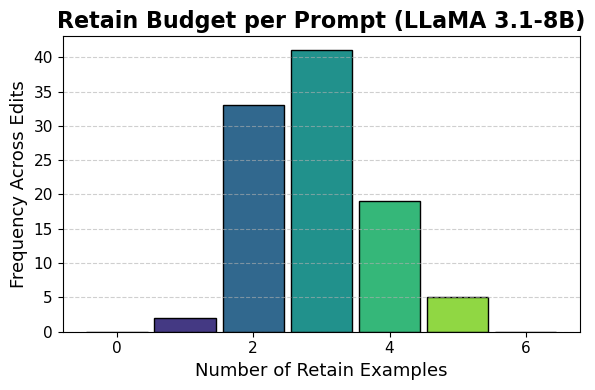}
  \caption{\textsc{Retain} budget distribution for LLaMA 3.1-8B under DR-IKE. Each bar shows the frequency of a given \textsc{Retain} count per prompt. Mean = $\mathbf{3.02}$, Std = $\mathbf{0.96}$.}
  \label{fig:llama_retain_budget}

  \vspace{1em}  

  \includegraphics[width=0.8\linewidth]{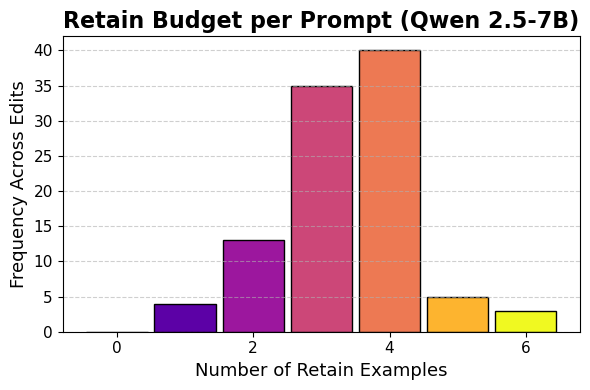}
  \caption{\textsc{Retain} budget distribution for Qwen 2.5-7B under DR-IKE. The model allocates more \textsc{Retain} examples for edits requiring stronger contextual support. Mean = $\mathbf{3.72}$, Std = $\mathbf{1.15}$.}
  \label{fig:qwen_retain_budget}
\end{figure}

Our proposed framework, \textbf{DR-IKE}, dynamically selects the number of \textsc{Retain} examples per prompt based on their estimated contribution to edit success, rather than applying a fixed number across all edits. Fig.~\ref{fig:llama_retain_budget} and Fig.~\ref{fig:qwen_retain_budget} illustrate how the number of \textsc{Retain} examples varies per edit instance for two different LLMs. In contrast to IKE, which statically includes 16 \textsc{Retains}  regardless of their informativeness or utility, DR-IKE tailors the prompt length to the needs of each specific edit.

This adaptivity yields significant efficiency gains. On average, LLaMA 3.1-8B requires only $\mathbf{3.02 \pm 0.96}$ \textsc{Retain}  examples per prompt, while Qwen 2.5-7B uses $\mathbf{3.72 \pm 1.15}$. These compact prompts reduce both the computational overhead and the risk of introducing irrelevant or distracting context. Notably, despite using fewer examples, DR-IKE maintains competitive or even superior edit success compared to IKE. This highlights the importance of selective inclusion: by pruning redundant or low-impact examples, DR-IKE not only saves prompt space but also enhances factual precision and generalization. The results underscore the utility of adaptive demonstration selection as a scalable solution for black-box knowledge editing.

\section{Appendix: Analysis of Budget Controller Update Strategies}
\setlength{\tabcolsep}{12pt}
\begin{table}[htb]
  \centering
  \begin{tabular}{|l|c|c|}
    \hline
    \textbf{Architecture} & \textbf{ESR ↑} & \textbf{PC ↑} \\ 
    \hline
    \makecell[l]{Current method \\ (baseline)} & \textbf{0.86} & 0.71 \\ 
    \hline
    \makecell[l]{MLP on softmax \\ scores} & 0.81 & \textbf{0.75} \\ 
    \hline
    \makecell[l]{MLP on softmax \\ + query features} & 0.81 & 0.69 \\ 
    \hline
    \makecell[l]{BERT + MLP \\ (predict $\sigma$)} & 0.83 & 0.65 \\ 
    \hline
    \makecell[l]{BERT \\(query + context) \\ + MLP} & 0.80 & 0.67 \\ 
    \hline
  \end{tabular}
  \caption{Comparison of alternative budget controller update strategies. 
  The current method achieves the best overall balance of Edit Success Rate (ESR) and Paraphrase Consistency (PC), 
  while other methods show improvements in isolated metrics but fail to consistently outperform the baseline.}
  \label{tab:budget_control}
\end{table}
We experimented with several alternative strategies for updating the budget controller, including semantic relevance modeling using BERT-based encoders, statistical modeling of the softmax score distribution, and hybrid approaches combining both. However, none of these alternatives consistently outperformed our current dynamic threshold update rule. In practice, we observed that the retriever dynamically adjusts its output probabilities over time. As $\sigma$ increases, the retriever tends to assign higher scores to borderline examples, leading to a natural stabilization of the threshold. This feedback mechanism allows $\sigma$ to converge without the need for complex or hand-crafted update rules.

Table~\ref{tab:budget_control} summarizes the different budget controller architectures we tested using the \textsc{LLaMA-3.2-3B} model, under the same train–test split as in our previous experiments. 
The \textbf{current method (baseline)} relies on a dynamic threshold update rule based on the retriever’s evolving score distribution. 
The \textbf{MLP on softmax scores} variant uses a lightweight feed-forward network trained directly on the softmax probability distribution, 
while \textbf{MLP on softmax + query features} extends this by incorporating query-level features. 
In the \textbf{BERT + MLP (predict $\sigma$)} approach, the query is first encoded with a BERT encoder and the resulting representation is passed to an MLP to directly predict the threshold $\sigma$. 
Finally, \textbf{BERT (query + context) + MLP} encodes both the query and the in-context examples with BERT before passing them to the MLP for threshold prediction.

Overall, while some alternatives improve on a single dimension (e.g., higher PC with softmax-based MLPs), none consistently match the balanced performance of the baseline approach. This confirms the effectiveness of our dynamic threshold update rule, though confidence-based calibration remains a promising direction for future refinement.

\section{Appendix: Sensitivity of Reward Function}

We tested the robustness of \textsc{DR-IKE} to input ordering and reward timing. 
Random shuffling or reordering of context examples left model outputs stable in the majority of cases. 
We also introduced random delays in the reward signal during training, and observed that Edit Success Rate (ESR) remained effectively unchanged, as summarized in Table~\ref{tab:reward_sensitivity}. 
These results indicate that the framework trains robustly even when the LLM’s binary reward signal is noisy or unstable.

\setlength{\tabcolsep}{12pt}
\begin{table}[htb]
  \centering
  \begin{tabular}{|l|c|}
    \hline
    \textbf{Sigma update method} & \textbf{ESR ↑} \\ 
    \hline
    No delay            & 0.7567 \\ 
    \hline
    Randomly delayed    & 0.7533 \\ 
    \hline
  \end{tabular}
  \caption{Sensitivity analysis of the reward function. Edit Success Rate (ESR) remains stable even with randomly delayed reward signals.}
  \label{tab:reward_sensitivity}
\end{table}

\section{Appendix: Additional Evaluation on Temporal and Numerical Facts}

Beyond the main benchmarks, we also conducted preliminary evaluation of \textsc{DR-IKE} on two additional settings. 
First, on a temporal-edit dataset provided by~\cite{zheng2023ike}, \textsc{DR-IKE} achieved a small improvement in Edit Success Rate (ESR) compared to the in-context baseline (\textsc{IKE}). 
Paraphrase Consistency (PC) could not be assessed in this case, as no paraphrased prompts were available in the dataset. 
Second, we extracted a small subset of numerical examples from the \textsc{CounterFact} dataset. 
On this subset, \textsc{DR-IKE} demonstrated consistently stronger performance in both ESR and PC relative to \textsc{IKE}. 
Since there is currently no dedicated dataset focused exclusively on numerical edits, and the few numerical cases in existing benchmarks are insufficient for a systematic study, we defer a more thorough investigation of numerical editing to future work.

\setlength{\tabcolsep}{12pt}
\begin{table}[htb]
  \centering
  \begin{tabular}{|l|c|c|}
    \hline
    \textbf{Method} & \textbf{ESR ↑} & \textbf{PC ↑} \\ 
    \hline
    \multicolumn{3}{|l|}{\textbf{Temporal Editing}} \\ 
    \hline
    IKE     & 0.46 & -- \\ 
    \hline
    \textbf{DR-IKE} & \textbf{0.50} & -- \\ 
    \hline
    \multicolumn{3}{|l|}{\textbf{Numerical Subset of CounterFact}} \\ 
    \hline
    IKE     & 0.67 & 0.53 \\ 
    \hline
    \textbf{DR-IKE} & \textbf{0.78} & \textbf{0.68} \\ 
    \hline
  \end{tabular}
  \caption{Preliminary evaluation of \textsc{DR-IKE} on temporal and numerical edits. 
  Paraphrase Consistency (PC) could not be computed for the temporal dataset due to lack of paraphrases.}
  \label{tab:temporal_numerical}
\end{table}

\end{document}